\documentclass[letterpaper, 10 pt, conference]{ieeeconf}  

\IEEEoverridecommandlockouts                              
\title{\LARGE \bf Jointly Learnable Behavior and Trajectory Planning\\for Self-Driving Vehicles }
\author{Abbas Sadat${}^{*,1}$, Mengye Ren${}^{*,1,2}$, Andrei Pokrovsky${}^{3}$, Yen-Chen Lin${}^{4}$, Ersin Yumer${}^{1}$, Raquel Urtasun${}^{1,2}$
\thanks{* Equal contribution}
\thanks{$^{1}$Abbas Sadat, Mengye Ren, Ersin Yumer and Raquel Urtasun are with Uber Advanced Technologies Group, 661 University Avenue, Suite 720, Toronto, Ontario, Canada, M5G 1M1. Email:
{\tt\small \{asadat,mren3,yumer,urtasun\}@uber.com}.}
\thanks{$^{2}$Mengye Ren and Raquel Urtasun are also with University of Toronto.}
\thanks{$^{3}$Andrei Pokrovsky is with GraphCore. Work done at Uber.}
\thanks{$^{4}$Yen-Chen Lin is with Massachusetts Institute of Technology. Work done at Uber.}
}
\usepackage{algorithmicx}
\usepackage{multicol}
\usepackage[bookmarks=true]{hyperref}

\usepackage{times}
\usepackage{epsfig}
\usepackage{graphicx}
\usepackage{amsmath}
\usepackage{amssymb}
\usepackage{url}
\usepackage{pst-all}
\usepackage{bm}
\usepackage{algorithm}
\usepackage[noend]{algpseudocode}
\usepackage{bbm}
\usepackage{enumerate}
\usepackage{subcaption}
\usepackage{multirow}
\usepackage{hhline}

\DeclareMathOperator*{\argmin}{argmin}

\newcommand{\etal}{\mbox{\emph{et al.\ }}}
\newcommand{\ie}{\mbox{\emph{i.e.\ }}}
\newcommand{\eg}{\mbox{\emph{e.g.\ }}}
\newcommand{\etc}{\mbox{\emph{etc.\ }}}

\newcommand{\bx}{\mathbf{x}}
\newcommand{\mT}{\mathcal{T}}
\newcommand{\mB}{\mathcal{B}}

\newcommand{\mW}{\mathcal{W}}
\newcommand{\mD}{\mathcal{D}}
\newcommand{\mL}{\mathcal{L}}
\newcommand{\mLB}{\mathcal{L}_M}
\newcommand{\mLC}{\mathcal{L}_I}
\newcommand{\lambdaB}{\lambda_M}
\newcommand{\lambdaC}{\lambda_I}
\newcommand{\bgB}{\mathbf{g}_M}
\newcommand{\bgC}{\mathbf{g}_I}
\newcommand{\cost}{c}
\newcommand{\bcost}{\mathbf{c}}
\newcommand{\bw}{\mathbf{w}}
\newcommand{\bg}{\mathbf{g}}

\newcommand{\dkurv}{\dot{\kappa}}

\newcommand{\baseline}{B+M}
\newcommand{\jointinfer}{B+M +J}
\newcommand{\jointlearn}{B+M +J +I}
\newcommand{\tb}{\textbf}
\begin{document}

\maketitle
\thispagestyle{empty}
\pagestyle{empty}

\begin{abstract}

The motion planners used in self-driving vehicles need to generate trajectories that are safe, comfortable, and obey the traffic rules.
This is usually achieved by two modules: \textit{behavior planner}, which handles high-level decisions and produces a coarse trajectory, and \textit{trajectory planner} that generates 
a smooth, feasible trajectory for the duration of the planning horizon. These planners, however, are typically 
developed separately, and changes in the behavior planner might affect the trajectory planner in unexpected ways. Furthermore, the final trajectory outputted by the trajectory planner might differ significantly from the one generated by the behavior planner, as they do not share the same objective.
In this paper, we propose a jointly learnable behavior and trajectory planner.
Unlike most existing learnable
motion planners that address either only behavior planning, 
or use an uninterpretable neural network to represent the entire logic from sensors to driving commands, 
our approach features an interpretable cost function on top of perception, prediction and vehicle dynamics, 
and a joint learning algorithm that learns a shared cost function employed by 
our behavior and trajectory components.  Experiments on real-world self-driving data demonstrate
 that jointly learned planner performs significantly better in terms of both similarity to human
driving and other safety metrics, compared to baselines that do not adopt joint
behavior and trajectory learning.

\end{abstract}

\section{Introduction}

Modern motion planners used in today's self-driving vehicles (SDVs) are typically composed of two distinct modules. The
first module, referred to as {\it behavior planning}, is responsible for providing high-level
decisions given the output of  perception and prediction (\ie perception outputs extrapolated to
future timestamps). To name a few, examples of such decisions are lane changes, turns and yields at an intersection. The second module, referred to as {\it trajectory
planning},  takes the decision of the behavior planner and a coarse trajectory and produces a
smooth trajectory for the duration of the planning horizon (typically 5 to 10s into the future).
This is then passed to the control module to execute the maneuver.

The role of the behavior planner is to constrain the trajectory generation such that a high-level objective is achieved. Earlier planners based on simple rule-based behavior selection or finite state machines are unfortunately unable to handle  decision making in complex real-world  urban scenarios. 
Alternative approaches try to optimize a behavioral objective by using sequential A* search \cite{ajanovic2018search} or parallel sampling methods \cite{werling2010optimal}, while reasoning about traffic-rules and other actors. 
A popular approach to alleviate the gap  between behavior and trajectory planning is to restrict the motion of the SDV to a path (\eg, lane centerline) and find the velocity profile that optimizes the behavioral objective. 

Trajectory planning is typically formulated as an optimization problem where a low-level objective is optimized locally to satisfy both the high-level decisions as well as  the kinematics and dynamics constraints \cite{bender2015combinatorial}. Many different continuous solvers, such as iLQR and SQP, have been exploited to solve the optimization problem. An alternative approach is sampling, where  trajectories are generated in a local region defined by the high-level decisions,  and the trajectory with lowest cost is selected for execution. Searching through a spatio-temporal state-lattice representing  continuous trajectories is another popular approach. 

While great progress has been achieved in the  development of individual methods for either behavior or trajectory planners, little effort has been dedicated to jointly designing these two modules. 
As a consequence, they typically do not share the same cost function, and thus changes in the behavior planner can have negative effects on the heavily tuned trajectory planner. Furthermore, the gains and costs of these planners are mostly manually tuned.
As a result, motion planning engineers spend a very significant amount of their development time re-tuning and re-designing the planners given changes in the stack. 


\begin{figure}[t]
\centering
\vspace{-0.1in}
\includegraphics[width=\columnwidth,trim={0 3.5cm 10cm 0}]{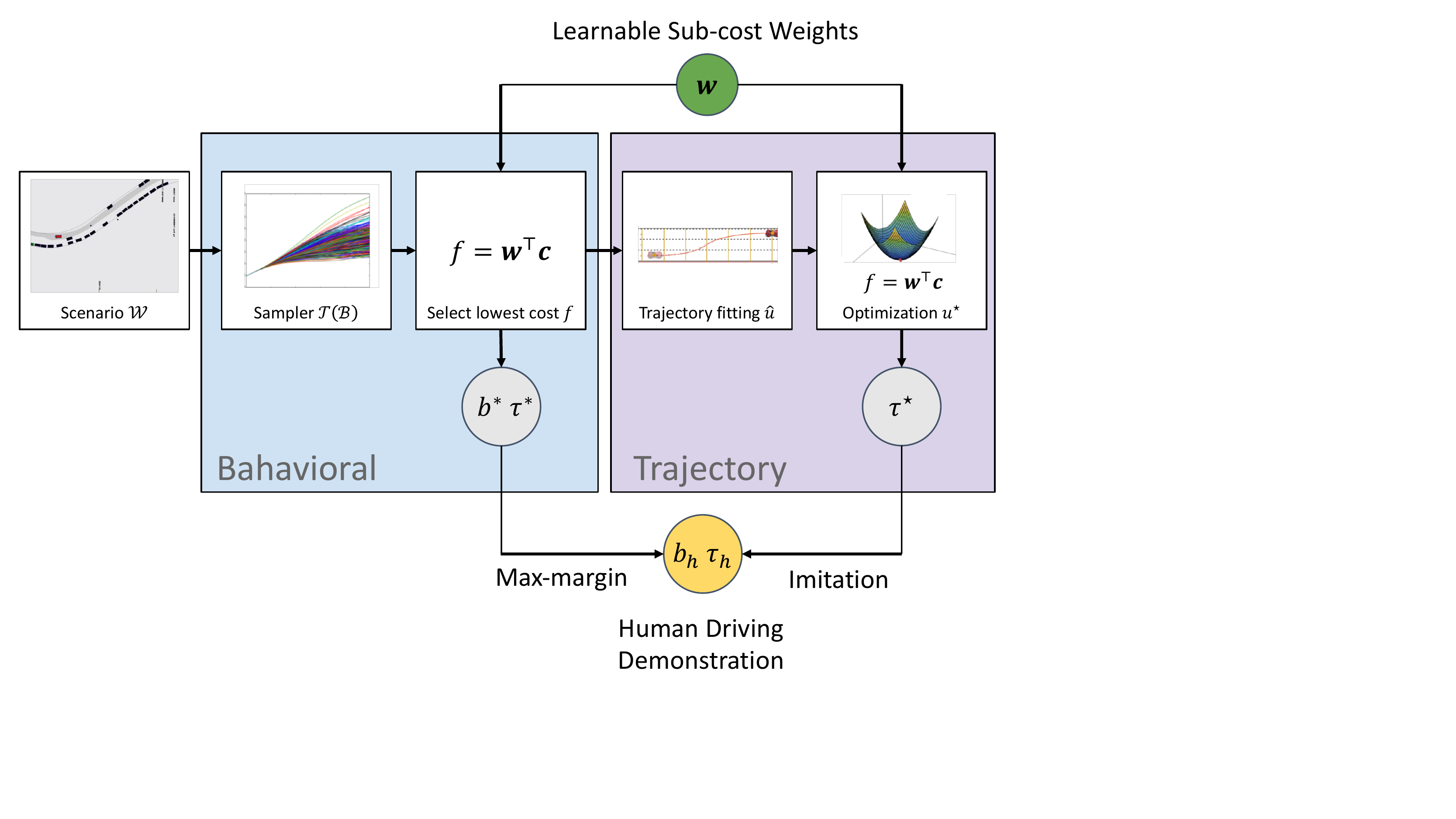}
\caption{Our learnable motion planner has discrete and continuous components, minimizing the same
cost function with a same set of learned cost weights. 
}
\label{fig:general}
\vspace{-0.2in}
\end{figure}

In this paper we tackle this issue by designing a motion planner where both the behavior and the trajectory planners share the same cost function. Importantly, our planner can be trained jointly end-to-end without requiring manual tuning of the costs functions.  
Our motion planner is designed to produce comfortable, safe, and mission-oriented trajectories and is able to handle complex traffic scenarios such as nudging to bicyclists and objects that partially occupy lanes, and yielding at intersections. 

We demonstrate the effectiveness of our approach in real-world urban scenarios by comparing the planned trajectories to comfortable trajectories performed by cautious human drivers. We show that our planner can learn to produce comfortable trajectories in terms of jerk and stay more close to the manually-driven trajectories when compared to  the baselines. Additionally, we demonstrate the performance of our planner on a very large-scale  dataset of highly challenging safety critical scenarios.

\section{Related Work}

\paragraph{Behavior Planning} 
Bender~\etal~\cite{bender2015combinatorial} propose to enumerate behaviors such that each discrete
class of behaviors could later be considered in an independent trajectory optimization step.
Gu~\etal~\cite{gu2016road} propose a multi-phase decision making framework where a traffic-free
reference planning over a long range is performed first, followed by a traffic-based refinement
where other actors are taken into account; lastly, a final step of local trajectory planning
addresses the short-term motion horizon in a refined manner. In \cite{ajanovic2018search},   a
search-based behavior planning approach that is capable of handling hundreds of variants in
real-time was proposed, addressing the limitation of~\cite{bender2015combinatorial}. This is
achieved by representing the search space and the driving constraints with a geometric
representation that is amenable to modeling predictive control schemes, and using an explicit
cost-to-go map. Recently, Fan~\etal~\cite{fan2018baidu} proposed a framework that uses dynamic
programming to find an approximate path and a speed profile iteratively in an EM-like  scheme,
followed by quadratic programming optimization of the cost function. In all of the above cases,
manually designed costs are used that consider lane
boundaries, collision, traffic lights, and other driving conditions. Even though hand tuning the
contribution of each constraint is possible, it is very time consuming.

\paragraph{Trajectory Planning}
Werling~\etal~\cite{werling2010optimal} introduced a combinatorial approach to trajectory planning, 
where a set of candidate swerve trajectories align the vehicle to the center of a lane given by an upstream
behavior planning.
Similarly, approaches with combinatorial schemes and
lattices were used in~\cite{Ziegler:2009:SSL:1733023.1733053},
\cite{Pivtoraiko:2009:DCM:1527169.1527172,mcnaughton2011motion}. Conversely, discretization is avoided in  \cite{ziegler2014trajectory}, by introducing a continuous non-linear
optimization method where obstacles in the form of polygons are converted to quadratic constraints.

\paragraph{Learned Motion Planning} 
Learning approaches to motion planning have mainly been studied from an imitation learning (IL)
~\cite{pomerleau1989alvinn,maxmarginplan,bojarski2016end,bansal2018chauffeurnet,codevilla2018end,muller2018driving,fan2018auto},
or  reinforcement learning
(RL)~\cite{pan2017virtual,paxton2017combining,kendall2018learning,rhinehart2018r2p2} perspective.
While most IL approaches provide an end-to-end training framework to control outputs from sensory
data, they suffer from compounding errors due to the sequential decision making process of
self-driving. Moreover, these approaches require significantly more data due to the size of
learnable parameters in modern networks.
Fan~\etal~\cite{fan2018baidu} propose a ranked based IL framework for learning the reward function
based on a linear combination of features. One major difference from our approach is that their
continuous optimizer is not jointly learned. The success of RL approaches to date has been  limited to only simulated environments or simple problems in robotics. More importantly, both IL
and RL approaches, in contrast to the  traditional motion planners,  are not interpretable. 
Recently, Zeng~\etal~\cite{zeng2019neural} introduced an end-to-end neural planner where sensor data is processed upto the end of a behavioral planner cost function, in a deep network together with perception and prediction outputs to increase interpretability of such approaches. Ratliff~\etal~\cite{maxmarginplan} use maximum margin for behavioral planning. In contrast, we develop a framework where we tackle both behavior planning and local trajectory planning with a shared cost function that can be learned end-to-end. 

The contribution of our work can be viewed as a combination of the advantages from both learned 
and traditional two-stage approaches:
1) Like the learned approaches, we help eliminate the time-consuming, error-prone, and iterative
hand-tuning of the gains of the planner costs. 
 2) Unlike the learned approaches above, we do so within a framework of  interpretable costs jointly
imposed on these modules. Therefore, even though our motion planner is data-driven, it still
uses the widely adapted, interpretable costing concepts for each driving constraint. Moreover, it is
also end-to-end trainable.
\section{Joint Behavior-Trajectory Planner}


Motion planners of modern self-driving cars are composed of  two modules. The  behavioral planner is responsible for making high level decisions. The trajectory planner takes the decision of the behavioral planner
and a coarse trajectory and produces a smooth trajectory for the duration of the planning horizon. Unfortunately these planners are typically developed separately, and changes in the behavioral planner might affect, in unexpected ways, the trajectory planner. Furthermore, the trajectory outputted by the trajectory planner might differ significantly in terms of behavior from the one returned by the behavioral planner as they do not share the same objective. 
To address this issue, in this paper we propose a novel motion planner where both the behavioral and trajectory planners share the same objective. 

\begin{figure*}
\centering
\vspace{0.08in}
\includegraphics[width=0.98\textwidth,trim={0.3cm 12.3cm 5cm 0.5cm},clip]{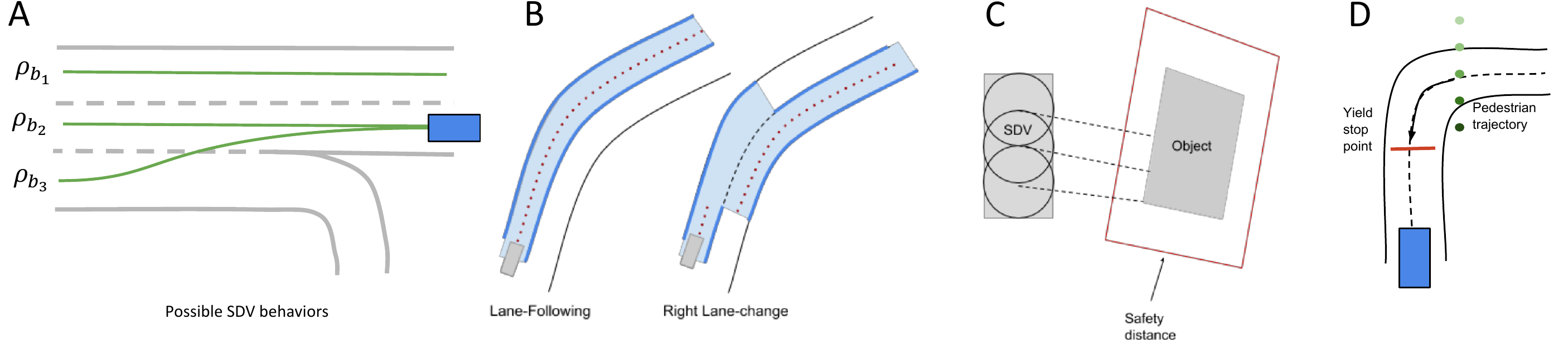}
\caption{A: Given a scenario, we generate a set of possible SDV behaviors. B: Left and right lane
boundaries and the driving path that are relevant to the intended behavior are considered in the
cost function. C: SDV geometry for spatiotemporal overlapping cost are approximated using circles.
D: The SDV yields to pedestrians through stop lines on the driving paths.}
\label{fig:multi1}
\end{figure*}

We use $\mW$ to denote the input to the motion planner from the upstream modules at each planning
iteration. In particular, $\mW$ includes the desired route as well as the state of the world, which contains the SDV state, the map, and the detected
objects. Additionally, for each object, multiple future trajectories are predicted including their probabilities.
The planner outputs a high-level behavior $b$
and a trajectory $\tau$ that can be executed by the SDV for the planning horizon $T=10s$. Here we
define
behavior as a driving-path that the SDV should ideally converge to and follow. 
These paths are obtained by considering \textit{keep-lane},
\textit{left-lane-change}, and \textit{right-lane-change} maneuvers. 
We refer the reader to Fig.~\ref{fig:multi1}-A for an illustration. 
At each planning iteration,
depending on the SDV location on the map, a subset of these behaviors, denoted by $\mB(\mW)$, is
allowed by traffic-rules and hence considered for evaluation. We  then generate low-level
realizations of the high-level behaviors by generating a set of trajectories $\mT(b)$ relative to
these paths (see Section~(\ref{sec:trajectory})). 
Assuming the SDV follows a bicycle model, we can
represent the vehicle state at time $t$ by $X_t=[\bx_t,
\theta_t,\kappa_t,v_t,a_t,\dkurv_t]$. Here $\bx$ is the Cartesian coordinate of position; $\theta$
is the heading angle; $\kappa$ is the curvature; $v$ is the velocity; $a$ is the acceleration; and
$\dkurv$ is the twist (derivative of curvature). A trajectory $\tau$ is defined as a sequence of
vehicle states at discrete time steps ahead.

The objective of the planner is then to find a behavior and a trajectory that is safe, comfortable, and
progressing along the route. We find such behavior and trajectory by minimizing a cost
function that describes the desired output:
\begin{align}
b^*, \tau^* = \argmin_{b\in\mB(\mW), \tau \in \mT(b)}  f(\tau, b,\mW; \bw)
\end{align}
We next describe the costs in more details, followed by our inference and learning algorithms.  

\section{A Unified Cost Function}
\label{sec:costs}

In this section we describe our unified cost function for our behavioral and trajectory planners.
Given the sets of candidate behaviors and trajectories, the cost function $f$ is used to choose
the best $(b,\tau)$. The cost function consists of sub-costs $\bcost$ that focus on different
aspects of the trajectories such as safety, comfort, feasibility, mission completion, and traffic
rules. We thus define
\begin{align}
f(\tau, b, \mW; \bw) = \bw^\top \bcost(\tau, b, \mW).
\end{align}
where the weight vector $\bw$ captures the importance of each sub-cost.
The following sub-sections introduce $\bcost$ in detail. 


\subsection{Obstacle}
A safe trajectory for the SDV should not only be collision-free, but also satisfy a safety-distance
to the surrounding obstacles, including both the static and dynamic objects such as vehicles,
pedestrians, cyclists, unknown objects, \etc 
Here we use $\cost_\text{overlap}$ and $\cost_\text{obstacle}$ to capture
the spatio-temporal overlap and violation of safety-distance respectively. For this, the SDV polygon is
approximated by a set of circles with the same radii along the vehicle, and we use the distance from
the center of the circles to the object polygon to evaluate the cost (see Fig.~\ref{fig:multi1}-C).
The overlap cost $\cost_\text{overlap}$ is then 1 if a trajectory
violates the spatial occupancy of any obstacle in a given predicted trajectory, and is averaged across
all possible predicted trajectories weighted by their probabilities.
The obstacle cost $\cost_\text{obstacle}$ penalizes the squared distance of the violation of the
safety-distance $d_\text{safe}$.
This cost is  scaled by the speed of the SDV, making the distance violation more
costly at higher speeds. This also prevents accumulating cost in a stopped trajectory when other
actors get too close to the SDV.


\subsection{Driving-path and lane boundary}
The SDV is expected to adhere to the structure of the road, \ie, it should not go out of the
lane boundary and should stay close to the center of the lane. Therefore, we introduce sub-costs
that measure such violations. The driving-path and boundaries that are considered for these
sub-costs depend on the candidate behavior (see Fig.~\ref{fig:multi1}-B).
The driving-path cost $\cost_\text{path}$ is the squared distance towards the driving path (red
dotted lines in Fig.~\ref{fig:multi1}-B). The lane boundary cost $\cost_\text{lane}$ is the
squared violation distance of a safety threshold. 


\subsection{Headway}
\label{sec:acc_cost}
As the SDV is driving behind a leading vehicle in either lane-following or lane-change behavior, it should
keep a safe longitudinal distance that depends on the speed of the SDV and the leading vehicle.
We compute the headway cost as the violation of the safety distance after
applying a comfortable constant deceleration, assuming that the leading vehicle applies a hard
brake \cite{shalev2017formal}.
\begin{figure}[t]
    \centering
\vspace{0.08in}
    \includegraphics[width=0.37\textwidth]{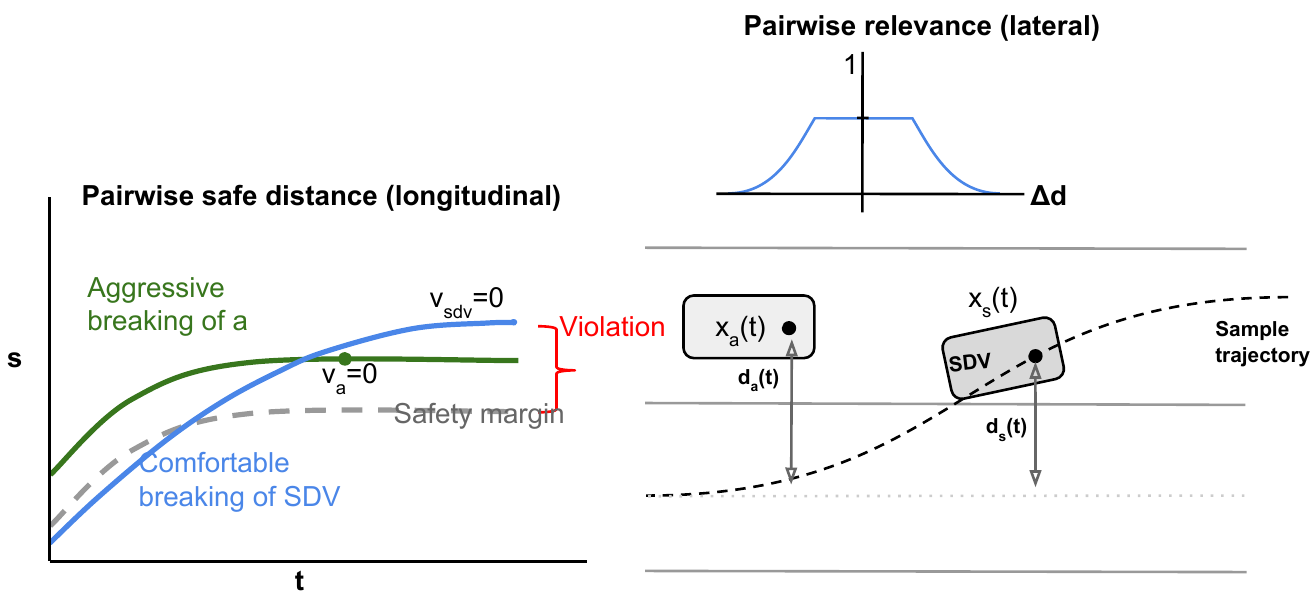}
    \caption{Left: Headway cost penalizes unsafe distance to leading vehicles. Right: for each sampled trajectory, a weight function determines how relevant an obstacle
    is to the SDV in terms of its lateral offset.}
    \label{fig:acc_cost}
    \vspace{-0.2in}
\end{figure}
To compute the cost above, we need to decide which vehicles are leading the SDV at each time-step in the planning horizon. A possible
approach is to associate vehicles to lanes based on 
distance to the center-line. However, this approach can be too conservative and make nudging
behavior difficult. Instead, we use a weight function of the lateral distance between the SDV and other vehicles to
determine how relevant they are for the headway cost (see Fig. \ref{fig:acc_cost}). Hence, the distance
violation costs incurred by vehicles that are laterally aligned with the SDV dominate the
cost. This is also compatible with lane change manoeuvres where deciding the lead vehicles can be
difficult. 

\subsection{Yield}
Pedestrians are vulnerable road users and hence require extra caution. When a pedestrian is
predicted to be close to the boundary of the SDV lane or crossing it, we impose a stopping point
at a safe longitudinal distance and penalize any trajectory that violates it (see
Figure~\ref{fig:multi1}-D). This is  different from a simple Cartesian distance as it
does not allow going around the pedestrians in order to progress in the route. The yield cost
$\cost_\text{yield}$ penalizes the squared longitudinal violation distance weighted by the
pedestrian prediction probability.
Similarly, the SDV needs to keep a safe longitudinal distance to vehicles that are predicted to be
crossing an intersection, as well as stop at signal-controlled intersections. We use the same cost
form as the pedestrian cost, but with different safety margins.

\subsection{Route}
The mission route is represented as a sequence of lanes, from which we can specify all lanes that
are on the route or are connected to the route by permitted lane-changes. A
behavior is desirable if the goal lane is closer to the route than the current lane. Therefore, we penalize the number of lane-changes that is required to converge to the route. 
Furthermore,
violation of a distance-threshold to the end of a lane is penalized to force lane-changes from
dead-end lanes to lanes that the SDV can continue on the route.


\subsection{Cost-to-go}
The sub-costs introduced so far evaluate the trajectory within the planning horizon, ignoring
what comes beyond it. Additionally, we incorporate a cost-to-go function to capture the value of the
final state of the SDV in a trajectory. This can prevent the planner to choose actions that are
sub-optimal beyond the horizon or, worst, take the SDV into an inevitable unsafe situation. 
For this purpose, we compute the deceleration needed for slowing-down to possible up-comming speed-limits and
use the square of the violation of the comfortable deceleration as cost-to-go. Consequently, trajectories
that end with high velocity close to turns or stop-signs  will be penalized.


\subsection{Speed limit, travel distance and dynamics}
\label{sec:other_costs}
Using the  speed-limit of a lane, which is available in the map data, we introduce a cost that
penalizes a trajectory if it goes above the eligible speed. The speed limit cost
$\cost_\text{speed}$ is the squared violation in speed.
In order to favor trajectories that advance in the route, we use the
travelled longitudinal distance as a reward.
Since the SDV is physically limited to certain ranges of acceleration, curvature,
{\it etc},  we prune trajectories that violate such constraints. Additionally, we introduce costs that
penalize aggressive motions to promote comfortable driving. Specifically, the dynamics cost
$\cost_\text{dyn}$ consists of the squared values of
jerk and violation thereof, 
acceleration and violation thereof, 
lateral acceleration and violation thereof, 
lateral jerk and violation thereof, 
curvature, 
twist,
and wrench.


\section{Inference}
\label{sec:inference}
\begin{algorithm}[t!]\small
    \caption{Inference of our joint planner}
    \label{alg:inference}
    \begin{algorithmic}[1] 
        \Procedure{Inference}{$\bw, \mW$}
            \Statex \Comment{The behavioral planner}
            \State $\tau^*, b^* \gets \argmin_{b \in \mB, \tau \in \mT(b)} f(\tau, b, \mW; \bw) $
            \State $u \gets$ \Call{TrajectoryFitter}{$\tau^*, b^*$}
            \Statex \Comment{The trajectory planner}
            \While{$u$ not converge}
                \State $u \gets$ \Call{OptimizerStep}{$f(\tau^{(T)}(u), b^*,\mW;\bw)$}
            \EndWhile
            \State $u^\star \gets u$
            \State $\tau^\star \gets \tau^{(T)}(u^\star)$
            \State \textbf{return} $\tau^\star, u^\star$
        \EndProcedure
    \end{algorithmic}
\end{algorithm}

In this section, we describe how our planner obtains the desired behavior and trajectory.
As shown in Algorithm~\ref{alg:inference} our inference process contains two stages of
optimization. In the
behavioral planning stage, we adopt a coarse-level parameterization for trajectory generation.
 The resulting trajectory is found by selecting the one with the lowest cost. 
In the trajectory planning stage, we use a fine-level parameterization where we model the
trajectory as a function of vehicle control variables. The trajectory is initialized with the
output of the behavior planner, and optimized through a continuous optimization solver.


\subsection{Behavioral Planner}
\label{sec:param}
We represent a trajectory in terms of the Frenet Frame of the driving-path of candidate
behaviors \cite{werling2010optimal}. Let $\Gamma_\rho$ be  the transformation from a bicycle model
state to the Frenet frame of a path $\rho$:
\begin{align}
[s,\dot{s},\ddot{s}, d, d^{\prime}, d^{\prime\prime}] = \Gamma_\rho(X),
\end{align}
where $s$ is the position (arc length) along the path, $d$ is the lateral offset.
$\dot{(.)}:=\frac{\partial}{\partial t}$, and $(.)^\prime:=\frac{\partial}{\partial s}$ denote the derivatives with respect to time and arc-length. Note that
the longitudinal state is parametrized by time, but the lateral state is parametrized by the
longitudinal position which, according to \cite{werling2010optimal}, is a better representation of
the coupling between the two states at relatively low speed. Figure~\ref{fig:gen_traj} sketches
the set of possible trajectories that can generate the nudging behavior of passing a vehicle.
\label{sec:trajectory}
Given an initial vehicle state $\Gamma_{\rho_b}(X_0)$, we generate longitudinal and lateral
trajectories as follows:

\subsubsection{Logitudinal trajectories}
The set of longitudinal trajectories $\mathcal{S}=\{s(t)\}$ are generated by computing an
exhaustive set of mid-conditions $[\dot{s}(t_1),t_1]$ and end-conditions $[\dot{s}(T),T]$ and
solving for two quartic polynomials stitched together. The acceleration ($\ddot{s}$) at $t_1$ and
$T$ are fixed at 0.

\subsubsection{Lateral trajectories}
Given a set of longitudinal trajectories, we parameterize lateral trajecotries $[d(s), d^\prime(s),
d^{\prime\prime}(s)]$ in terms of the longitudinal distance $s$. We generate a set of mid-conditions
$[d(s_1),s_1]$ and  fix $d^\prime(s_1)$ and $d^{\prime\prime}(s_1)$ to be 0. We also fix the
end-conditions to be $[0, 0, 0]$ so that the SDV is merged to the driving path. We stitch two
quintic polynomials to fit the mid- and end-conditions.

\begin{figure}[t]
\centering
\includegraphics[width=0.3\textwidth]{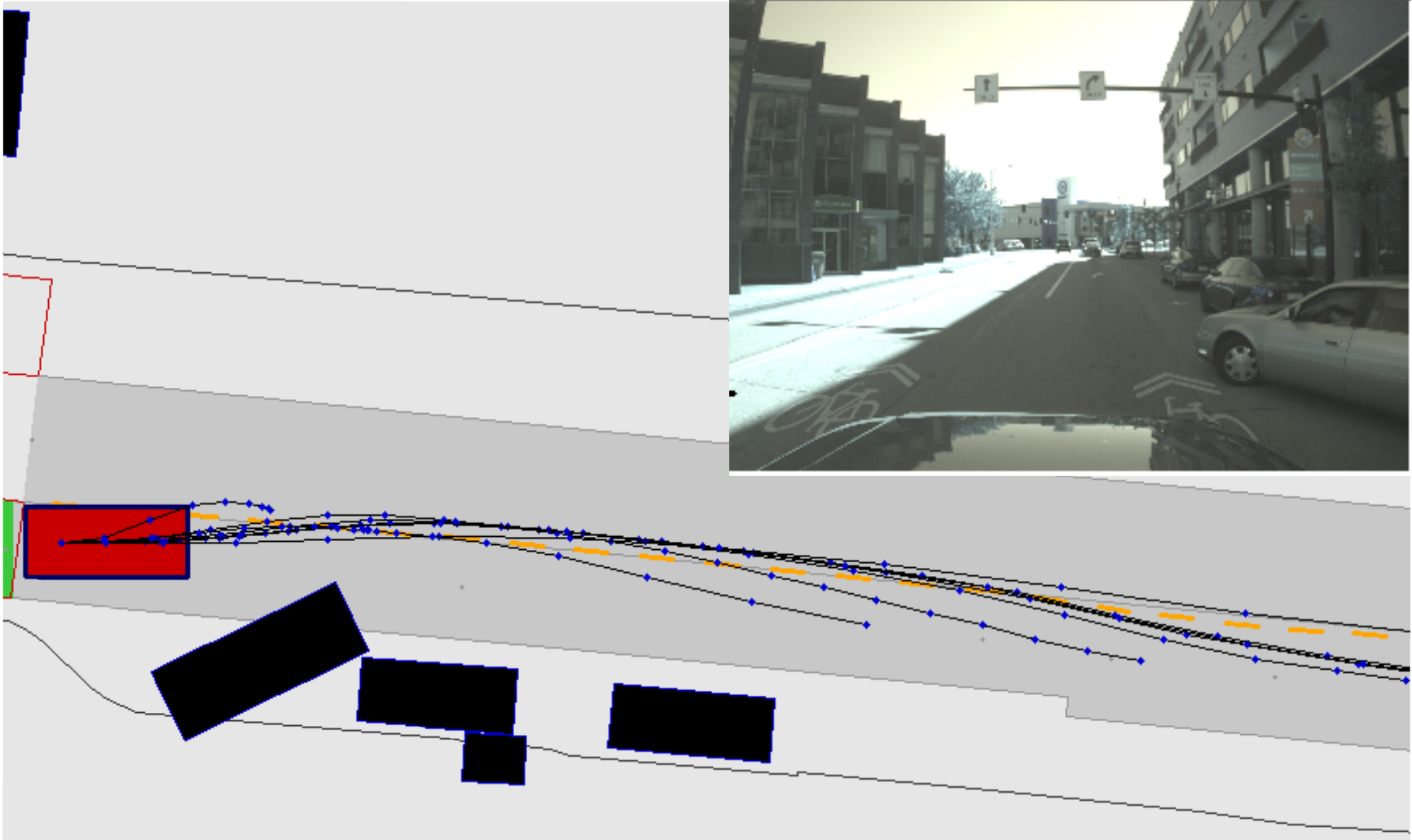}
\caption{Example trajectories in a nudging scenario.}
\label{fig:gen_traj}
\vspace{-0.05in}
\end{figure}

For computing the dynamics cost in the discrete planner, we transform each pair
of sampled longitudinal and lateral trajectories $[s(t),d(s)]$ back to a bicycle model trajectory:
\begin{align}
\tau &= [\bx,\theta,\kappa,v,a,\dot{\kappa}] = 
\Gamma_{\rho}^{-1}(s,\dot{s},\ddot{s}, d, d^{\prime}, d^{\prime\prime}).
\end{align}
\vskip -3mm
\looseness=-1
Figure \ref{fig:gen_traj} shows an example set of generated bicycle model trajectories.  
The optimal trajectory for a given scenario $\mW$ is found by evaluating the cost function $f$
for all $b \in \mB(\mW)$ and $\tau \in \mT(b)$, and choosing the one that achieves the minimum cost.

\subsection{Behavioral-Trajectory Interface}
\begin{figure}[t]
\centering
\includegraphics[width=0.3\textwidth,trim={0cm 11cm 21cm 0cm},clip]{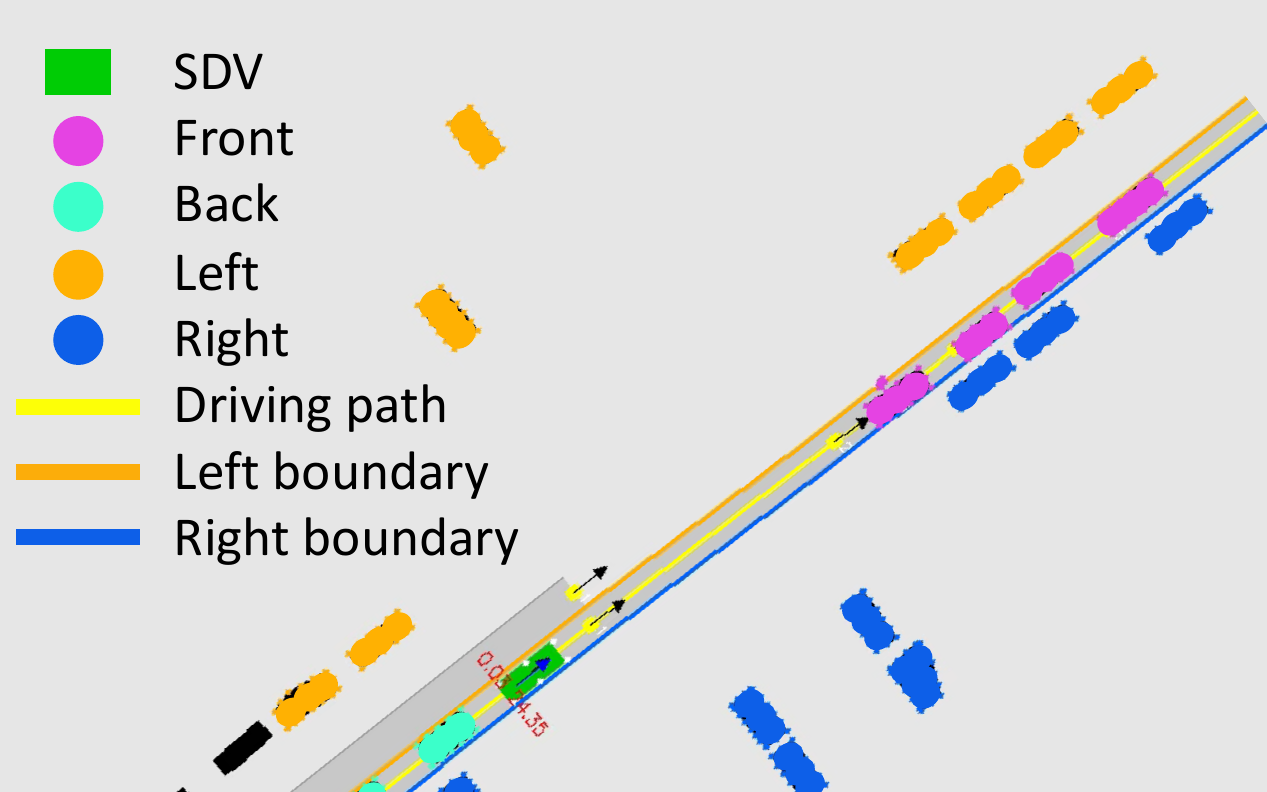}
\caption{Behavioral decisions include obstacle side assignment and lane information, which are sent
through the behavioral-trajectory interface. 
\vspace{-0.2in}
}
\label{fig:interface}
\end{figure}
The behavioral-trajectory interface passes the optimal behavioral decision $b^*$ and coarse-level
trajectory $\tau^*$ to the trajectory planning stage. $b^*$ is encoded as the left and right lane
boundaries, the driving path, and the obstacle side assignment, which determines whether an obstacle
stays in the front, back, left, or right to the SDV at time step $t$ (see Fig.~\ref{fig:interface}).
For the trajectory planner, spatio-temporal overlap cost $\cost_\text{overlap}$ will be incurred
if the side assignment is violated at any time in the planning horizon, scaled by the squared
distance of violation. This encourages that the trajectory planner respects the discrete decision
made by the discrete stage.


\vspace{-0.1in}
\subsection{Trajectory Planner}
The behavioral planner uses finite differences to estimate the control parameters, which may not
be precise for long range. 
Therefore, we use a trajectory fitter to compute the control parameters.
In this stage, we represent trajectories in the Cartesian coordinates: $\tau = (\mathbf{x}, \theta, v, a,
\kappa,
\dot{\kappa})$. 
We parameterize the trajectory
$\tau$ using the control variables jerk $j$ and wrench $\ddot{\kappa}$ (second derivative of
curvature). We use bicycle dynamics $\mD$ to model $\tau$ as a function of the controls.
\vskip -8mm
\begin{align}
\tau_{t} &= \mD(\tau_{t-1}, u_{t}),\\
\tau &= \{\mD(\tau_{t-1}, u_t)\}_{t=1}^T = \tau^{(T)}(u).
\end{align}
\vskip -3mm
Trajectory fitting minimizes the following objective wrt. the control variables:
\vskip -8mm
\begin{align}
\hat{u} &= \argmin_{u} \sum_{t=1}^{T} c_\bx(\bx(u))_t +
\lambda_\theta \sum_{t=1}^T c_\theta(u)_t + \lambda_\text{dyn} \cost_{\text{dyn}},
\end{align}
\vskip -3mm
where $c_\bx$ is the squared Euclidean distance between the trajectory positions, $c_\theta$ is
the orientation difference:
\vskip -8mm
\begin{align}
c_\theta(u)_t=\frac{1}{2}\left\lVert \left(
\begin{array}{c}
\sin\theta(u)_t \\
\cos\theta(u)_t
\end{array}\right) -
\left(
\begin{array}{c}
\sin\hat\theta_t \\
\cos\hat\theta_t
\end{array}\right) \right\rVert^2_2,
\end{align}
\vskip -3mm
and $\cost_{\text{dyn}}$ is the set of costs related to vehicle dynamics described in
Section~\ref{sec:other_costs}. This allows us to start the optimization process with a physically
feasible trajectory.

Given a fitted control sequence as initialization, the continuous optimization module achieves a
local minimum of the overall cost function $f$:
\vskip -8mm
\begin{align}
u^\star = \argmin_{u} f(\tau^{(T)}(u), b^*, \mW; \bw).
\end{align}
\vskip -5mm
We utilize the BFGS solver to obtain the solution of the above optimization problem.
\vspace{-0.1in}

\vspace{-0.1in}
\section{Learning}
\vspace{-0.1in}
\label{sec:learning}
 We use a combination of max-margin objective and imitation learning as our loss function
\vskip -8mm
\begin{align}
\label{eq:maxmargin}
\vspace{-0.1in}
\mL(\bw) = \frac{\lambda_\bw}{2}& \lVert{\bw}\rVert^{2}_2 + \lambdaB \mLB(\bw) + \lambdaC \mLC(\bw),
\end{align}
\vskip -3mm
where $\mLB$ is the max-margin loss,  $\mLC$ is the imitation learning loss, and $\lambdaB$ and
$\lambdaC$ are hyperparameters that scale the two loss components.

\begin{minipage}[t!]{0.95\linewidth}
\centering
\vspace{-0.1in}
\begin{algorithm}[H]\small
    \caption{Learning of our joint planner}
    \label{alg:learning}
    \begin{algorithmic}[1] 
        \Procedure{Learning}{$\bw^{(0)}$}
            \For{$i \gets 1...N$}
                \State $E_M \gets \Call{GetMiniBatch}{}$, $\bgB \gets 0$
                \State $E_I \gets \Call{GetMiniBatch}{}$, $\bgC \gets 0$
                \Statex \Comment{Max-margin learning}
                \For{$E_j = (\tau_h, b_h, \mW) \in E_M$}
                    \For{$b \in \mB, \tau \in \mT(b)$}
                    \State $l_D(b, \tau) \gets \Delta(\tau_{h}, b_h, \tau, b) - f(\tau, b,\mW; \bw^{(i-1)})$
                    \EndFor
                    \State $\{b^*_k, \tau^*_k\} \gets \text{top-K}\ l_D(b, \tau)$ s.t. $b \in \mB, \tau \in \mT(b)$
                    \State $\bgB \gets \bgB + \frac{1}{K|E_D|}  \sum\limits_{k} (\bcost_\mW(\tau_h, b_h)-\bcost_\mW(\tau^*_k, b^*_k))$
                \EndFor\\
                \Statex \Comment{Differentiable inference imitation learning}
                \For{$E_j = (\tau_h, b_h, \mW) \in E_I$}
                    \State $u^\star_0 \gets$ \Call{Optimize}{$f(\tau^{(T)}(u), b^*, \mW; \bw)$}
                    \Statex \Comment{Gradient descent $M$ steps}
                    \For{$m \gets 1...M$}\label{line:gd_start}
                        \State $u^\star_m \gets$ $u^\star_{m-1} - \eta \nabla_u f(u^\star_m)$
                    \EndFor\label{line:gd_end}

                    \Statex \Comment{Backprop through time (BPTT)}
                    \State $l_C \gets \frac{1}{T}\sum_t \gamma^t \lVert \bx^\star_t - \bx_{h,t} \rVert_2^2 $
                    \For{$m \gets M...1$}\label{line:bptt_start}
                        \State $\bgC \gets \bgC + \frac{1}{|E_I|}\nabla_u l_C(u^\star_m) \nabla_\bw u^\star_m$
                    \EndFor
                \EndFor

                \State $\bg \gets \lambda_\bw \bw^{(i-1)} + \lambdaB \bgB + \lambdaC \bgC$
                \Comment{Sum up gradients}
                \State $\bw^{(i)} \gets \bw^{(i-1)} \exp(-\alpha \bg)$
                \Comment{Exp. gradient descent} 
            \EndFor

            \State \textbf{return} $\bw^{(N)}$
        \EndProcedure
    \end{algorithmic}
\end{algorithm}
\vspace{0.1in}
\end{minipage}

The max-margin learning loss penalizes trajectories that have small cost and are different from
the human driving trajectory \cite{maxmarginplan}. Let $\{(\mW,\tau_h, b_h)\}_{i=1}^N$ be a set of
manual-driving examples where $b_h$ and $\tau_h$ denote the
ground-truth human behavior and trajectory respectively. We learn the linear weights $\bw$ of the
cost function $f$ using  structured SVM. This encourages the human driving trajectory to have
smaller cost than other trajectories. In particular,
\vspace{-0.1in}
\begin{eqnarray}
\label{eq:maxmargin}
\label{eq:surrogate}
\mLB(\bw) = \frac{1}{N} \sum_{i=1}^N \bigg\{ f(\tau_{h,i}, b_{h,i}, \mW_i; \bw) + \nonumber\\
\max\limits_{b\in\mathcal{B}(\mW_i), \tau \in
\mathcal{T}(b)} \left\{\Delta(\tau_{h,i}, b_{h,i}, \tau, b) -f(\tau, b, \mW_i; \bw)\right\} \bigg\},
\end{eqnarray}
and $\Delta(\tau_h, b_h, \tau, b)$ is the task-loss, which measures the dissimilarity between
pairs of $(\tau, b)$. It consists of the L1 distance between the positions of the trajectories,
and constant offsets for any behavioral differences and undesirable outcomes. The maximization can
be solved by treating the task-loss as a sub-cost in $f$. In practice, we select the top $K$
trajectories that have the maximal values.

For the imitation loss function $\mLC$, we use {\it mean square error} (MSE) for measuring the
distance between positions of the human trajectory and the planner optimal trajectory, with some
discounting factor $\gamma$:
\begin{align}
\mLC(\bw) = \frac{1}{2N} \sum_{i=1}^N \sum_{t=1}^T \gamma^t \lVert \bx^\star_t - \bx_{h,t}
\rVert^2_2
\end{align}

The overall gradient of the learning objective can be written as $\bg = \lambda_\bw \bw + \lambdaB
\bgB + \lambdaB \bgC$, where $\bgC = \nabla_\bw \mLC$ and $\bgB$ is the subgradient of $\mLB$:
\begin{align}
\bgB = \frac{1}{NK}\sum_{i=1}^N\sum_{k=1}^K \bcost(\tau_{h,i}, b_{h,i},
\mW_i)-\bcost(\tau^*_{i,k}, b^*_{i,k}, \mW_i),
\end{align}
where $\{\tau^*_k, b^*_k\}$ are $K$ maximum violation examples and $\bcost$ is the sub-cost vector.

The max-margin objective uses a surrogate loss to learn the sub-cost weights, since selecting the
optimal trajectory within a discrete set is not differentiable. In contrast, the iterative
optimization in the trajectory planner is a differentiable module, where gradients of the
imitation loss function can be computed using the backpropagation through time (BPTT) algorithm
\cite{bptt}. Since unrolling the full optimization can be computationally expensive, we unroll
only for a truncated number of steps after we obtain a solution. As shown in
Line~\ref{line:gd_start} in Alg.~\ref{alg:learning}, we perform $M$ gradient descent steps after
obtaining the optimal trajectory, and backpropagate through these $M$ steps only
(Line~\ref{line:bptt_start}). If the control obtained from the continuous optimization converges
to the optimum, then backpropagating through a truncated number of steps is approximating of the
inverse Hessian ($[\nabla^2_{uu}f]^{-1}$) at the optimum $u^\star$.

Since we would like the weights to be greater than zero, we use the exponentiated gradient descent
update on the sub-cost weights:
$\bw^{(i+1)} = \bw^{(i)} \exp(-\alpha \bg)$,
where $\alpha$ is the learning rate parameter. This update ensures that the learned weights are
always positive.


\section{Experiments}
\begin{table*}[ht!]
\centering
\vspace{0.08in}
\resizebox{0.7\linewidth}{!}{
\begin{tabular}{|l|c c c| c c c c c|}
\hline
\multirow{2}{*}{Method} & \multicolumn{3}{ c| }{$L_2$ (m)} & \multicolumn{5}{ c |}{Driving Metrics}      \\
       & 1.0s & 2.0s & 3.0s & Jerk & Lat. accel. & Speed lim. & Progress & Diff. behavior (\%)           \\
\hline                                                                                             
\hline                                                                                             
HUMAN    &-          &-          &-          &0.237      & 0.168       &0.14       &3.58      &-         \\
\hline\hline                                                                                             
Oracle                                                                                                   
ACC      &0.700      &1.560      &3.670      &-          &-            &0.34       &3.64       &-        \\ 
PT       &0.254      &0.941      &2.190      &0.147      &\tb{0.232}   &\tb{0.26} &\tb{3.81}  &8.5      \\
Ours                                                                                                     
$\bw$    &0.095      &0.712      &1.992      &0.140      &0.264        &0.39       &\tb{3.81}  &\tb{0.0} \\
Ours                                                                                                     
$\bw_b$  &0.094      &0.699      &1.945      &\tb{0.135} &0.234        &0.27       &3.72       &\tb{0.0} \\
Ours                                                                                                      
$\bw_b^t$&\tb{0.066} &\tb{0.514} &\tb{1.495} &0.143      &0.279        &0.28       & 3.78      &\tb{0.0} \\
\hline
\end{tabular}
}
\caption{Human $L_2$ and driving metrics on ManualDrive}
\label{tab:l2}
\vspace{-0.2in}
\end{table*}

We conducted our 
experiments in {\it open-loop simulation}, where we unroll the planner independent of the new
observations.

\subsection{Datasets}
We use two real-world driving datasets in our experiments. \textbf{ManualDrive} is a set of human driving recordings where the drivers are instructed to drive
   smoothly and carefully respecting all traffic rules. There are
\{12000, 1000, 1600\} scenarios in the training, validation and test set respectively. We use
this dataset for training and evaluation on major metrics. 
Since this dataset does not contain labeled objects, we additionally use 
\textbf{TOR-4D}, which is composed of very challenging scenarios.  We exploit the labeled objects in 3D space to compute spatiotemporal overlap metrics. We use this dataset for evaluation only. In particular our test set contains 5500 scenarios.

\subsection{Evaluation metrics}
We use several metrics for evaluation. 

\paragraph{Similarity to human driving} We use the average $\ell_2$ distance between the planned
trajectories and human trajectories within \{1.0, 2.0, 3.0\} seconds into the future. Lower number means that the
planner is behaving similarly to a human driver. 

\paragraph{Passenger comfort} We measure passenger comfort in terms of the average jerk and lateral
acceleration.

\paragraph{Spatiotemporal overlap} A good planning should avoid obstacles. We measure the
spatiotemporal overlap with obstacles within \{1.0, 2.0, 3.0\} seconds into the future, in terms of the
percentage of  scenarios. We use a perception and prediction (P\&P) module as input
to the planner. We also report the percentage of overlap excluding the obstacles that are not 
present in detection or predicted not to have overlap with the SDV, including vehicles that are more than
25m behind the SDV as they are assumed to react to the ego vehicle (referred to as ``excl. non-overlap pred.").
\begin{table}[t]
\vspace{0.08in}
\centering
\begin{tabular}{|l|c c c|c c c|c c c|l}
\hline
\multirow{2}{*}{Method} & \multicolumn{3}{ c| }{All} & \multicolumn{3}{ c |}{excl. non-overlap pred.}\\
                & 1.0s     & 2.0s     & 3.0s      & 1.0s     & 2.0s     & 3.0s \\
\hline                                                                               
\hline                                                                               
Oracal ACC      & 0.780    & 1.300         & 2.990     & -         & -        & -        \\
PT              & 0.290    & 1.480         & 3.230     & 0.036     & 0.670    & 2.130    \\
Ours $\bw$      &\tb{0.000}&\tb{0.128}     & 1.010     &\tb{0.000} &0.055     & 0.600    \\
Ours $\bw_b$    &\tb{0.000}&\tb{0.128}     & 1.010     &\tb{0.000} &0.055     & 0.560    \\
Ours $\bw_b^t$  &\tb{0.000}&\tb{0.128}     & \tb{0.826}&\tb{0.000} &\tb{0.036}&\tb{0.340}\\
\hline
\end{tabular}
\caption{Spatiotemporal overlapping rate (\%) on TOR-4D}
\label{tab:tor4d}
\end{table}

\paragraph{Others} We also measure other aspects of driving: average violation of speed limit
(km/h),  progress in the route, and proportion of the scenarios where the planner chooses a
different behavior compared to the human. 

\subsection{Baselines}
We use \textbf{Oracle ACC} as baseline.  Note that this baseline uses the ground-truth driving
path, and follows the leading vehicle's driving trajectory. We call this oracle as it has access
to the ground-truth behavior. Our second baseline, called  {\bf PT},  is a simpler version of our
behavioral planner. It only reasons in the longitudinal-time space, without the lateral dimension.
In lane change decisions, it projects obstacles from both lanes onto the same longitudinal path.
The weights are also learned using the max-margin formulation described in
Section~\ref{sec:learning}

\subsection{Experiment setup}
As described  in \ref{sec:costs}, we have 30 different sub-costs. 
We consider the following {\bf model variants} with increasing complexity:
\begin{itemize}
    \item $\bw$ learns a single weight vector for all scenarios. This model learns 30 different weights.
    \item $\bw_b$ learns a separate weight vector for each behavior: keep lane, left lane change, and
    right lane change. Thus it has $3\times 30 = 90$ weights for 3 behaviors. 
    \item $\bw_b^t$ learns a separate weight vector for each behavior and at each time step. This
    variant can automatically learn planning cost discounting factor for the future. This model has $3 \times 30 \times 21 =
1890$ weights for 21 timesteps. 
\end{itemize}

The following {\bf ablation variants} are designed to validate the usefulness of our proposed joint
inference and learning procedure.
\begin{itemize}
    \item {\it Behavioral with max-margin (``\baseline'')} learns the weight vector through the
    max-margin (+M) learning on the behavioral planner only.
    \item {\it Full Inference (``\jointinfer'')} uses the trained weights of ``\baseline'', and
    runs the joint inference algorithm (+J) at test time.
    \item {\it Full Learning \& Inference (``\jointlearn'')} learns the weight vector using the
    combination of max-margin (+M) and imitation objective (+I), and runs the joint inference
    algorithm (+J) at test time.
\end{itemize}

{\bf Training details:}
We pretrain the behavioral planner with the max-margin objective for a fixed number of steps, and
then start joint training of the full loss function. For fair comparison, the baseline (B+M) is
also trained for the same number of steps. For learning the imitation loss during training, we use
human trajectory controls as the initialization for the trajectory optimization for training
stability.

\begin{table*}
\vspace{0.08in}
\centering
\resizebox{0.7\linewidth}{!}{
\begin{tabular}{|l|c c c| c c c c|}
\hline
\multirow{2}{*}{Method} & \multicolumn{3}{ c| }{$L_2$ (m)} & \multicolumn{4}{ c |}{Driving Metrics}  \\
       & 1.0s & 2.0s & 3.0s & Jerk & Lat. accel. & Speed lim. & Progress \\ 
\hline\hline
$\bw$ \baseline  
       &0.241      &0.900      &2.100      & 0.143     &0.257   &\tb{0.22} &3.72     \\
$\bw$ \jointinfer
       &0.096      &0.727      &2.030      & 0.141     &\tb{0.239}        & 0.39     &3.80      \\
$\bw$ \jointlearn
       &\tb{0.095} &\tb{0.712} &\tb{1.992} &\tb{0.140} &0.264        &0.39      &\tb{3.81}  \\
\hline\hline
$\bw_b$ \baseline  
       &0.241      &0.890      &2.070      & 0.145     &0.282   &\tb{0.23} &3.73      \\
$\bw_b$ \jointinfer
       &0.097      &0.720      &2.000      & 0.143     &0.264        & 0.37     &\tb{3.79} \\
$\bw_b$ \jointlearn
       &\tb{0.094} &\tb{0.699} &\tb{1.945} &\tb{0.135} &\tb{0.234}        &0.27      &3.72     \\
\hline\hline
$\bw_b^t$ \baseline  
       &0.240      &0.790      &1.750      & \tb{0.136}     &0.330   &\tb{0.20} &3.72      \\
$\bw_b^t$ \jointinfer
       &\tb{0.066}      &\tb{0.514}      &1.501      & 0.143     &\tb{0.278}       & 0.29     &\tb{3.80} \\
$\bw_b^t$ \jointlearn
       &\tb{0.066} &\tb{0.514} &\tb{1.495} &0.143 &0.279        &0.28      &3.78      \\
\hline
\end{tabular}
}
\caption{Effect of joint learning and inference on ManualDrive}
\label{tab:ablation_l2}
\vspace{-0.2in}
\end{table*}

\subsection{Results and Discussion}
Table~\ref{tab:l2} presents our results on the ManualDrive dataset. The best model $\bw_b^t$
clearly outperforms other variants and baselines. In terms of similarity to human trajectories
within 3.0s time horizon, the full model has a relative improvement of 31.7\% over the PT baseline
which uses a simpler cost function without a trajectory planner, and 24.9\% over the simplest
model variant $\bw$. Although the full model is still a linear combination of all the sub-costs,
$\bw_b^t$ has almost 2000 free parameters and it would be virtually impossible to manually tune
the coefficients. Our jointly learned models also shows better behavioral decision compared to PT. 
We also provide other driving metrics in the table; however, they are
less indicative of the model performance. For example, the human driving trajectories are reported
to have a higher jerk and make less distance progress.

Table~\ref{tab:tor4d} presents our results on the TOR-4D dataset, where we measure the
spatiotemporal overlapping of our SDV with other obstacles. Our jointly learned model shows lower
chance of overlapping with obstacles: compared to the PT baselines, our best model reduced 73\% of
the spatiotemporal overlapping within 3.0s time horizon.

\textbf{Joint inference and learning:} Table~\ref{tab:ablation_l2},\ref{tab:tor4d_ablation}
presents results when the joint inference or learning modules are removed from our system. Shown
in Table~\ref{tab:ablation_l2}, the full model shows a clear improvement in terms of the $\ell_2$
position error compared to human. This is expected since it is optimized as the imitation loss.
Compared to the model that does trajectory planner only for inference, the jointly learned model
shows better performance in terms of jerk and lateral acceleration. In
Table~\ref{tab:tor4d_ablation}, the full model achieves the lowest spatiotemporal overlap rate.
This suggests that by treating the trajectory optimization as a learnable module, our planner
learns to produce safe and smooth trajectories by imitating human demonstrations.
\begin{table}
\centering
\resizebox{0.9\linewidth}{!}{
\begin{tabular}{|l|c c c|c c c|l}
\hline
\multirow{2}{*}{Method}& \multicolumn{3}{ c| }{All} & \multicolumn{3}{ c |}{excl. non-overlap pred.}\\
           & 1.0s  & 2.0s & 3.0s & 1.0s  & 2.0s & 3.0s \\
\hline
\hline
$\bw$ \baseline
&0.110     &0.330     & 1.390      &0.018     &0.091     & 0.690    \\
$\bw$ \jointinfer
&\tb{0.000}&\tb{0.110}& 1.020     &\tb{0.000}&\tb{0.055}&0.620     \\
$\bw$ \jointlearn
&\tb{0.000}&    0.128 &\tb{1.010} &\tb{0.000}&\tb{0.055}&\tb{0.600}\\
\hline\hline
$\bw_b$ \baseline
&0.011          &0.330     & 1.370     &0.018     &0.091     & 0.690    \\
$\bw_b$ \jointinfer
&\tb{0.000}     &\tb{0.128}& 1.040     &\tb{0.000}&\tb{0.055}& 0.580    \\
$\bw_b$ \jointlearn
&\tb{0.000}     &\tb{0.128}&\tb{1.010} &\tb{0.000}&\tb{0.055}&\tb{0.560}\\
\hline\hline
$\bw_b^t$ \baseline
&0.110     &0.360     & 1.130    &0.018     &0.091     &0.360      \\
$\bw_b^t$ \jointinfer
&\tb{0.000}&0.146     & 0.936    &\tb{0.000}&\tb{0.018}&0.360      \\
$\bw_b^t$ \jointlearn
&\tb{0.000}&\tb{0.128}&\tb{0.826}&\tb{0.000}&0.036     &\tb{0.340} \\
\hline
\end{tabular}
}
\caption{Spatiotemporal overlapping rate (\%) on TOR-4D}
\label{tab:tor4d_ablation}
\vspace{-0.2in}
\end{table}

\vspace{-0.4cm}
\section{Conclusion}

In this paper we proposed a learnable end-to-end behavior and trajectory motion planner. Unlike
most existing learnable motion planners that address either behavior planning, or use an
uninterpretable neural network to represent the entire logic from sensors to driving commands, our
approach features an interpretable cost function and a joint learning algorithm that learns a
shared cost function employed by our behavior and trajectory components. Our experiments on real
self-driving datasets demonstrate that the jointly learned planner performs significantly better
in terms of both similarity to human driving and other safety metrics. In the future, we plan to
explore utilizing our jointly learnable motion planner to train perception and prediction modules.
This is possible as   we can back-propagate through it. We expect this to significantly improve
these tasks.

\vspace{-0.4cm}

\bibliographystyle{unsrt}
\bibliography{refs}
\end{document}